\documentclass[]{spie}  %>>> use for US letter paper
\usepackage{amsmath,amsfonts,amssymb}
\usepackage{graphicx}
\usepackage[colorlinks=true, allcolors=blue]{hyperref}

\title{Image-based marker tracking and registration for intraoperative 3D image-guided interventions using augmented reality}

\author[a]{Andong Cao}
\author[b]{Ali Dhanaliwala}
\author[c]{Jianbo Shi}
\author[b]{Terence Gade}
\author[b]{Brian Park}
\affil[a]{Yale University, New Haven, CT USA 06520}
\affil[b]{Penn Image-Guided Interventions Lab, University of Pennsylvania, 646 Curie Blvd, Philadelphia, PA USA 19104}
\affil[c]{GRASP Lab, University of Pennsylvania, Philadelphia, PA USA 19104}

\begin{document} 
\maketitle

\begin{abstract}
Augmented reality (AR) has the potential to improve operating room workflow by allowing physicians to ``see" inside a patient through the projection of imaging directly onto the surgical field. For this to be useful the acquired imaging must be quickly and accurately registered with patient and the registration must be maintained. Here we describe a method for projecting a CT scan with Microsoft Hololens (Hololens) and then aligning that projection to a set of fiduciary markers. Radio-opaque stickers with unique QR-codes (markers) are placed on an object prior to acquiring a CT scan. The location of the markers in the CT scan are extracted and the CT scan is converted into a 3D surface object. The 3D object is then projected using the Hololens onto a table on which the same markers are placed. We designed an algorithm that aligns the markers on the 3D object with the markers on the table. To extract the markers and convert the CT into a 3D object took less than 5 seconds. To align three markers, it took $0.9 \pm 0.2$ seconds to achieve an accuracy of $5 \pm 2$ mm. These findings show that it is feasible to use a combined radio-opaque optical marker, placed on a patient prior to a CT scan, to subsequently align the acquired CT scan with the patient. 
\end{abstract}

\keywords{Augmented Reality, Mixed Reality, Interventional Radiology, Computer Vision, Registration}

\begin{figure} [ht]
   \begin{center}
   \begin{tabular}{c} %% tabular useful for creating an array of images 
   \includegraphics[width=0.7\textwidth]{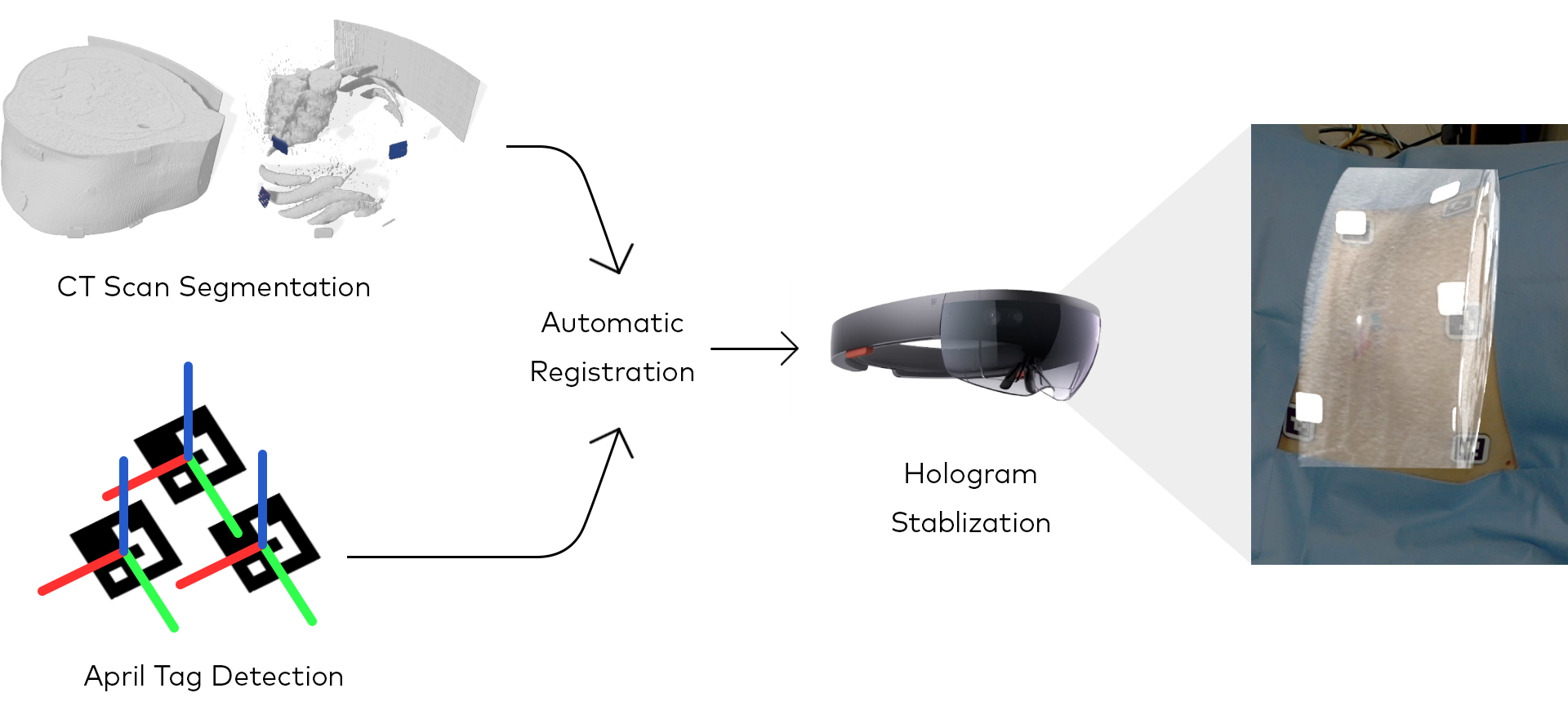}
   \end{tabular}
   \end{center}
   \caption[example] 
%>>>> use \label inside caption to get Fig. number with \ref{}
   { \label{fig:overview} 
Overview of the proposed workflow to perform automatic registration for AR-assisted surgical navigation.}
\end{figure}

\section{INTRODUCTION}
\label{sec:intro}  % \label{} allows reference to this section
Accurate registration and tracking are essential for any surgical navigation system. Recently, HoloLens has been shown to be systematically superior to comparative devices for surgical applications \cite{frantz2018augmenting, qian2017comparison}. Computer vision approaches have been used to improve anatomic 3D model tracking and registration, but most use custom image-based markers, such as QR codes or patterns, that are unable to be used in sterile environments or during procedures. Ideally, markers need to be unique and able to be easily detected by computer vision as well as easily detected on medical imaging. Visimarkers (Clear Guide Medical, Baltimore, MD)  are sterile, radiopaque markers with AprilTags that are used for image fusion and registration. To our knowledge, we do not know of any other available image-based markers that are radiopaque and easily detectable on CT scans as well as being sterile.

In this paper, we propose a workflow to use Visimarkers to perform automatic registration for AR-assisted surgical navigation. We developed a method to segment out optical markers on CT scans, and a fast and accurate algorithm for marker detection and registration. Compared to common registration methods, such as the Iterative Closest Point (ICP) algorithm \cite{horn1988closed, bentley1975multidimensional} and the Vuforia library (Vuforia), our algorithm can offer either faster registration speed or higher registration accuracy, while capable of performing registration with sparse and noisy point correspondences.

\section{Methods}
\label{sec:methods}

\subsection{CT Segmentation}

VisiMarkers were placed on a subject’s skin prior to a routine CT exam. The CT scan was then anonymized and exported as a DICOM image set. The Image set was then imported into GNU Octave \cite{GNU} where segmentation and analysis was performed. The segmentation procedure is depicted in Figure \ref{fig:overview}. To produce a 3D object for display with the Hololens, the DICOM image set was imported into the 3D viewer plugin of FIJI \cite{schindelin2012fiji} from which it was exported as an STL file followed by a final conversion into an OBJ file using Blender.

% \begin{figure} [ht]
%   \begin{center}
%   \begin{tabular}{c} %% tabular useful for creating an array of images 
%   \includegraphics[width=\textwidth]{img/Calibration2.png}
%   \end{tabular}
%   \end{center}
%   \caption[example] 
% %>>>> use \label inside caption to get Fig. number with \ref{}
%   { \label{fig:setup} 
% (a) The calibration setup, where the coordinate system is center at a marker, and the offset of the marker's hologram is set to a fixed value (e.g. 0.01 meters in the image). (b) Left: the plot of the sampled measurements and the contour plot of the fitted offset in the $x$-$y$ plane. Right: the plot of the sampled measurements and the contour plot of the fitted offset in the $y$-$z$ plane. Note that the circle around each measurement is the corresponding error bar.}
% \end{figure}

\subsection{Hololens Calibration}

The calibration process for the Hololens device involves two steps. First, following a standard camera calibration procedure developed for a pinhole camera \cite{camera_calibration}, the intrinsic parameters of the Hololens are estimated using a standard ChARUCO board \cite{charuco, charuco2}. Second, to correct the misalignment between the hologram and the corresponding marker, the position shift of the hologram is measured, modelled, and corrected. 

\begin{figure} [ht]
  \begin{center}
  \begin{tabular}{c} %% tabular useful for creating an array of images 
  \includegraphics[width=0.7\textwidth]{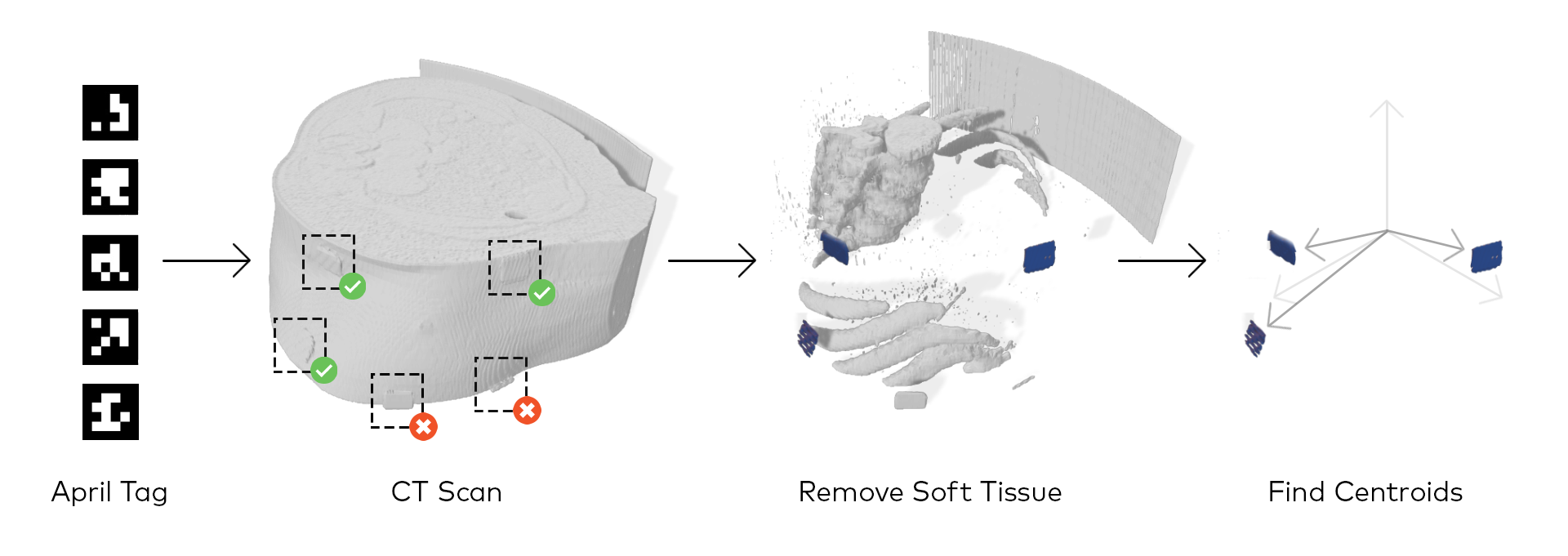}
  \end{tabular}
  \end{center}
  \caption[example] 
%>>>> use \label inside caption to get Fig. number with \ref{}
  { \label{fig:ctscan} 
The procedure to segment out the marker (April Tags). First, apply the markers onto the patient's body. Second, remove the soft tissue from the CT scan and identify connected components. Third, filter out objects of sizes different from the expected size of the marker, and calculate the centroid of the detected marker.}
\end{figure}

\subsection{Registration}
\begin{figure} [ht]
   \begin{center}
   \begin{tabular}{c} %% tabular useful for creating an array of images 
   \includegraphics[width=0.7\textwidth]{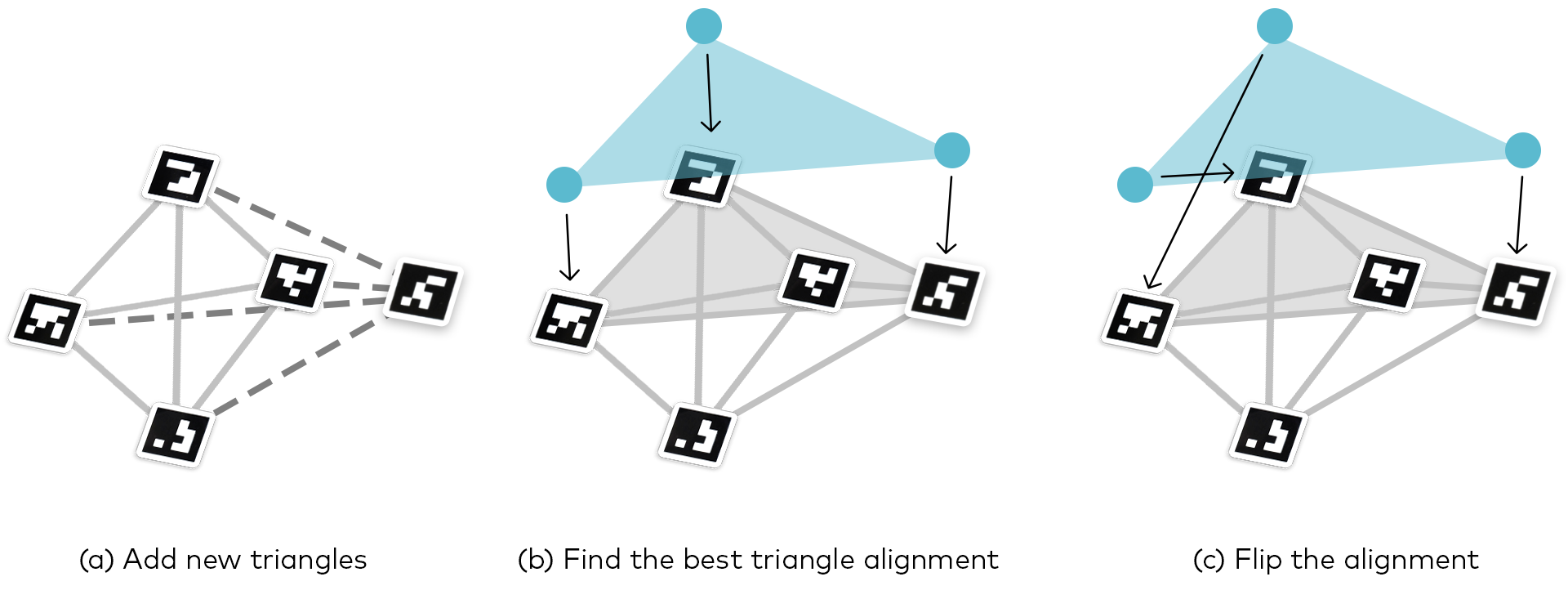}
   \end{tabular}
   \end{center}
   \caption[example] 
%>>>> use \label inside caption to get Fig. number with \ref{}
   { \label{fig:registration_algorithm} 
Overview of the registration algorithm. Left: when a new marker appears, record all new triangles that contains this marker. Middle: find the detected triangle that is most similar to the triangle on the CT scan, and align the corresponding triangles together. Right: if the registration results in a wrong normal, then flip the alignment by exchange one pair of the point correspondence}
\end{figure}

One of the main challenges of marker registration in a surgical setting is the limited number of available point correspondences. In such scenario, ICP and Vuforia become highly sensitive to noises in the position measurement. We developed an algorithm that is robust under sparse point correspondences while achieving high accuracy and registration speed. In the following experiments, we test the most extreme condition where there is only three detected tags on the CT scan. Our registration algorithm is based on aligning triangles formed by the segmented markers on the CT scan and those formed by the detected markers on the patient's body. This method keeps track of all the detected triangles, finds the one that is most similar to the triangle on the CT scan, and align the corresponding triangles together. Each triangle is encoded in a $k$-d tree \cite{bentley1975multidimensional} sorted by the ratio of the three edges $e_1:e_2:e_3$ of the triangle, where $e_1$ is the longest edge and $e_2$ is the shortest edge. Figure \ref{fig:registration_algorithm} illustrates the main steps of the algorithm.

\section{Results}
\label{sec:results}

\subsection{Registration Speed}

We compared our registration algorithm with two other popular methods, Vuforia and ICP. The registration times for our algorithm and the ICP is evaluated using the CPU utilization rate in the Windows Device Portal connected to the Hololens. First, we measured the CPU utilization rate during idle app runtime for one minute, and found that the average CPU utilization rate is about $87 \pm 3\%$. Second, we ran the registration algorithm, and recorded the elpased time during which the CPU utilization rate stays above the average rate measured previously. The rest of the registration speed data, including the registration speed of Vuforia, is referenced from the paper by Park et. al. \cite{park2019registration}

\begin{figure} [ht]
   \begin{center}
   \begin{tabular}{c} %% tabular useful for creating an array of images 
   \includegraphics[width=0.8\textwidth]{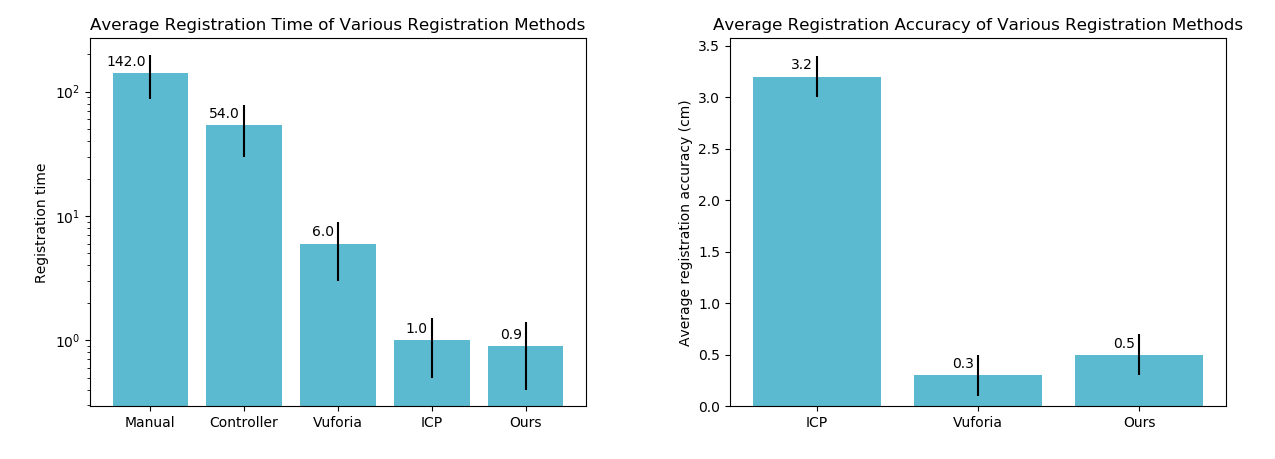}
   \end{tabular}
   \end{center}
   \caption[example] 
%>>>> use \label inside caption to get Fig. number with \ref{}
   { \label{fig:registration_time} 
Left: average registration time of various registration methods. Right: average registration accuracy of various registration methods.}
\end{figure}

\begin{figure} [ht]
   \begin{center}
   \begin{tabular}{c} %% tabular useful for creating an array of images 
   \includegraphics[width=0.7\textwidth]{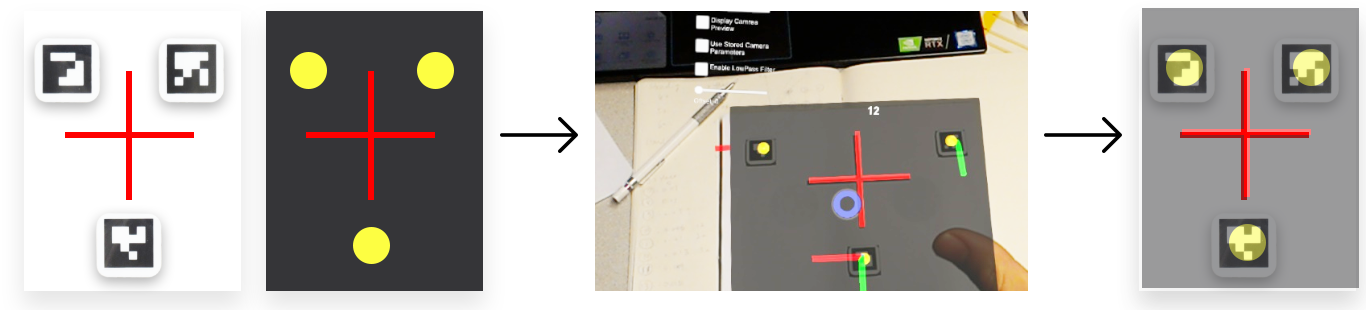}
   \end{tabular}
   \end{center}
   \caption[example] 
%>>>> use \label inside caption to get Fig. number with \ref{}
   { \label{fig:registration_method} 
   Left: real reference image and the hologram of the reference image. Middle: mark the center of the hologram on the real reference image. Right: measure the distance between the marked center and the actual center of the reference image.
}
\end{figure}

\subsection{Registration Accuracy}
\begin{figure} [ht]
   \begin{center}
   \begin{tabular}{c} %% tabular useful for creating an array of images 
   \includegraphics[width=\textwidth]{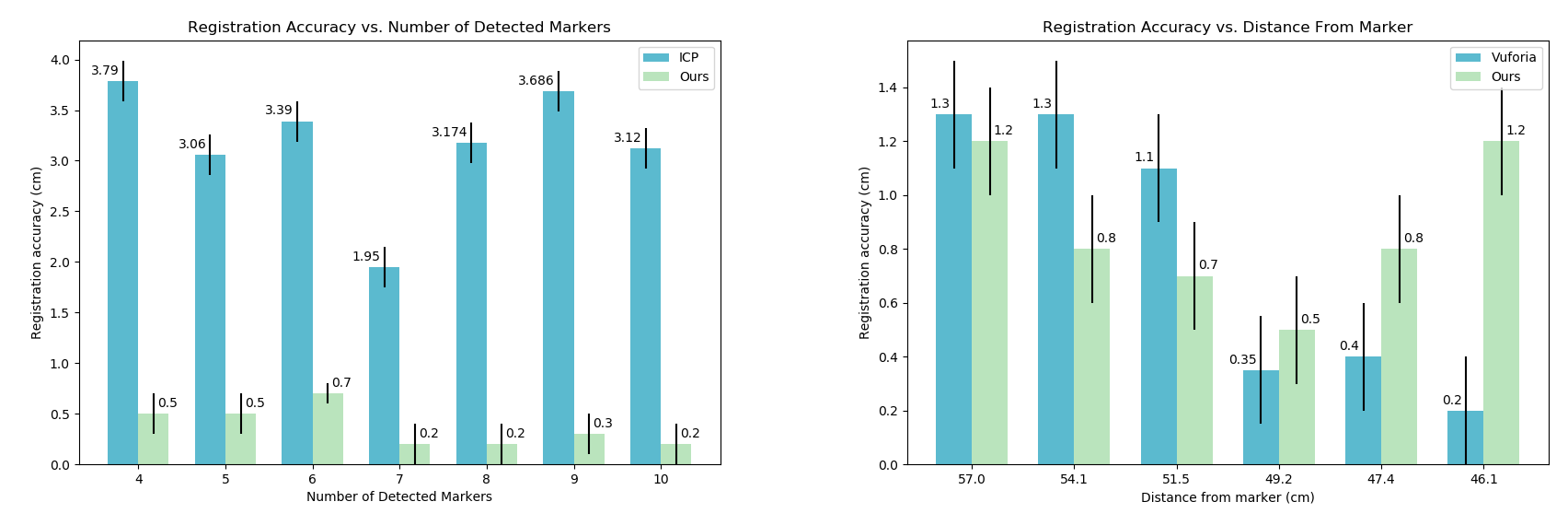}
   \end{tabular}
   \end{center}
   \caption[example] 
%>>>> use \label inside caption to get Fig. number with \ref{}
   { \label{fig:series_accuracy} 
Left: registration accuracy vs. number of detected markers. Right: registration accuracy vs. distance from marker.}
\end{figure}

The registration accuracy is measured by tracing the center of the hologram of a reference image overlaying a real copy of that image, and measuring the difference between the center of the two images, as shown in Figure \ref{fig:registration_method}. The registration accuracy of Vuforia is measured via the same method but with a different marker, namely the one employed by Park et. al. \cite{park2019registration}, because the reference image used by our algorithm contains too little feature for Vuforia to track. The left subplot of Figure \ref{fig:series_accuracy} shows that, for ICP and our algorithm, the registration accuracy depends on the number of detected markers. Note that the registration accuracy of our algorithm decreases from $0.5 \pm 0.2$ cm to $0.2 \pm 0.2$ cm as the number of markers increases from 4 to 10. The right subplot of Figure \ref{fig:series_accuracy} compares the performance of Vuforia and our algorithm under distance variations. In order to evaluate the effect of distance on registration accuracy, we placed the markers at six different distances from the Hololens, and measured the registration accuracy using both Vuforia and our algorithm. The data suggest that Vuforia has a shorter detection range than our algorithm, and its performance becomes better as the distance decreases. On the other hand, our algorithm’s registration accuracy reaches a local minimum at distance of 51.5 cm. This distance is approximately one arm’s length away from the tag, allowing a natural pose for the surgeon in an operating room.

% \subsection{Number of Iterations}

% When the initial point correspondence is incorrect, most registration algorithms will run another iteration to start from a different point alignment. The number of corrective iterations indicates the efficiency of the registration algorithm. This metric only applies to ICP and our algorithm, because Vuforia does not show its internal iterations during registration processes.
% As the number of detected markers increases, the number of possible point alignments increases and the number of iterations increases accordingly. Figure depicts the comparison between the number of iterations of our algorithm and that of the ICP, with respect to the number of detected tags.

\section{New or Breakthrough Work}
Our goal is to implement a complete pipeline to align a patient's CT scan with the patient in real-time. In this report, we describe specifically the segment of our pipeline for registering cross sectional imaging data back onto a patient. The key insight here is the utilization of radio-opaque markers. By placing these markers on the patient before a scan, fiduciary markers become embedded directly into the image data, providing us with two sets of points to be aligned. Using these point clouds, we can now produce accurate registration rapidly since we can eliminate the time consuming step of manual refinement of the registration that was previously required to achieve our designated level of accuracy.

\section{Conclusions}

We developed a method that can accurately use radio-opaque optical marker to perform CT scan registration, and evaluated its registration speed and accuracy compared to common alternatives such as Vuforia and ICP algorithms. This method can be applied in a real-world surgical field, potentially allowing physicians to “see” inside a patient to significantly improve procedure time.
\newpage
% References
\bibliography{report} % bibliography data in report.bib
\bibliographystyle{spiebib} % makes bibtex use spiebib.bst

\end{document}